\documentclass{article}
\usepackage{spconf,amsmath,graphicx}
\usepackage{color}
\usepackage{multirow}
\usepackage{amssymb}
\usepackage{bm}
\usepackage[breaklinks=true,bookmarks=false,colorlinks=True]{hyperref}

\def\eg{{\em e.g.}}

\title{Fast Portrait Segmentation with highly light-weight network}
\name{Yuezun Li$^{1*}$, Ao Luo$^{2*}$\thanks{$*$ indicates equal contribution.} and Siwei Lyu$^1$}
\address{$^1$ University at Albany, State University of New York, USA \\ $^2$ University of Electronic Science and Technology of China}

\begin{document}
%
\maketitle
\begin{abstract}
In this paper, we describe a fast and light-weight portrait segmentation method based on a new {\em highly light-weight backbone} (HLB) architecture. The core element of HLB is a {\em bottleneck-based factorized block} (BFB) that has much fewer parameters than existing alternatives while keeping good learning capacity. Consequently, the HLB-based portrait segmentation method can run faster than the existing methods yet retaining the competitive accuracy performance with state-of-the-arts. Experiments conducted on two benchmark datasets demonstrate the effectiveness and efficiency of our method.
\end{abstract}

\vspace{-0.3cm}
\section{Introduction}
\label{sec:intro}
\vspace{-0.3cm}
The proliferation of smart phones and social portals (\eg, {\tt Facebook}, {\tt Twitter}) has made capturing and sharing personal photos ever more convenient. Many mobile photo editing software is based on the function of {\em portrait segmentation}, which is the task to separate the region corresponding to the subject's head and upper-body from a selfie photo for subsequent operations such as background replacement, hair style change, and non-photo realistic rendering of the portrait.

\begin{figure}[t]
    \centering
    \includegraphics[width=\linewidth]{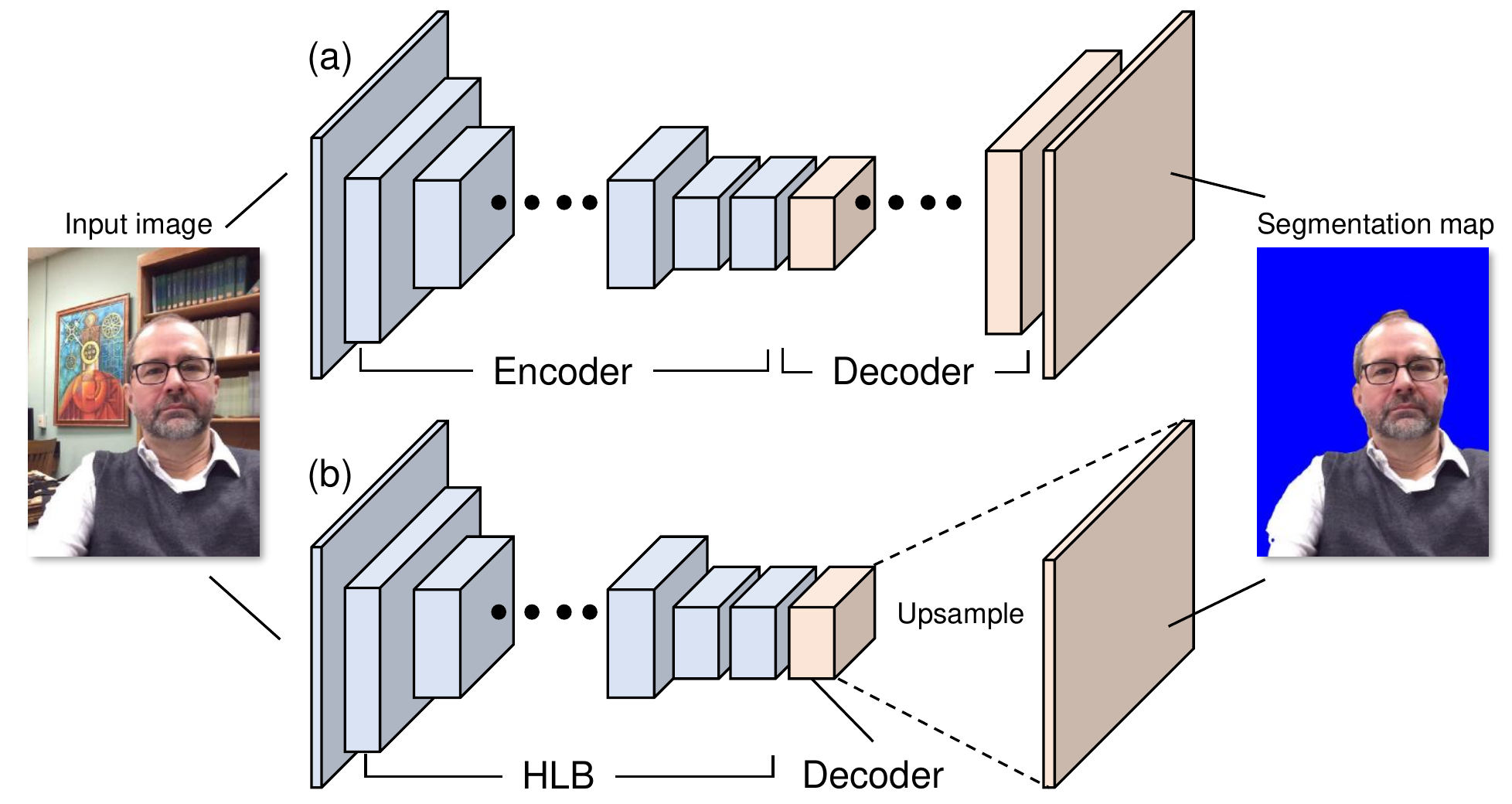}
    \vspace{-0.8cm}
    \caption{\em \small (a) is the general architecture of portrait segmentation. Given an input image, the encoder extracts the features, which are sent to decoder for segmentation map prediction. (b) is HLB-based portrait segmentation, where the decoder only contains one convolutional layer.}
    \label{fig:teaser}
    \vspace{-0.5cm}
\end{figure}

Currently, the most effective portrait segmentation methods are based on convolutional neural networks (CNN), \eg,  \cite{shen2016automatic,shen2017high,du2019boundary}. These methods adopt a general architecture composed of an encoder (backbone) and a decoder network, see Fig.\ref{fig:teaser}(a). 
To achieve good performance, these methods usually employ powerful CNN models as VGG \cite{simonyan2014very} and ResNet \cite{he2016deep}. However, such full-blown CNN models may not scale up well in running efficiency when performing portrait segmentation over large number of images and/or on a lean computing platform such as cellphones. For such tasks, we need a fast, light-weight yet powerful portrait segmentation method. Several more recent works, \eg, \cite{zhang2019portraitnet,zhu2019portrait}, use existing general purpose light-weight network structures such as the MobileNet \cite{howard2017mobilenets} to improve the overall running efficiency. 

In this paper, we describe a fast and light-weight portrait segmentation method based on a new {\em highly light-weight backbone} (HLB) architecture.
The core element of this backbone architecture is a novel residual block, which we term as the {\em bottleneck-based factorized block} (BFB), combining the bottleneck \cite{he2016deep} and factorized layers \cite{alvarez2016decomposeme}. The bottleneck structure is composed by two $1 \times 1$ convolutional layers at the beginning and the ending of block, which reduces the width of feature maps inside the block. The factorized layer replaces the standard 2D convolution kernels with two 1D convolution kernels to reduce the number of parameters. 
The existing methods usually employ multiple layers as decoder to generate the favorable results. In contrast, due to the good learning capacity of HLB architecture, we can achieve the competitive results by only using a single convolutional layer as the decoder (Fig.\ref{fig:teaser}(b)).
Experiments conducted on two benchmark datasets demonstrate the effectiveness and efficiency of our method.

\vspace{-0.7cm}
\section{HLB Architecture}
\label{subsec:backbone_arch}
\vspace{-0.3cm}

\begin{figure}[t]
    \centering
    \includegraphics[width=\linewidth]{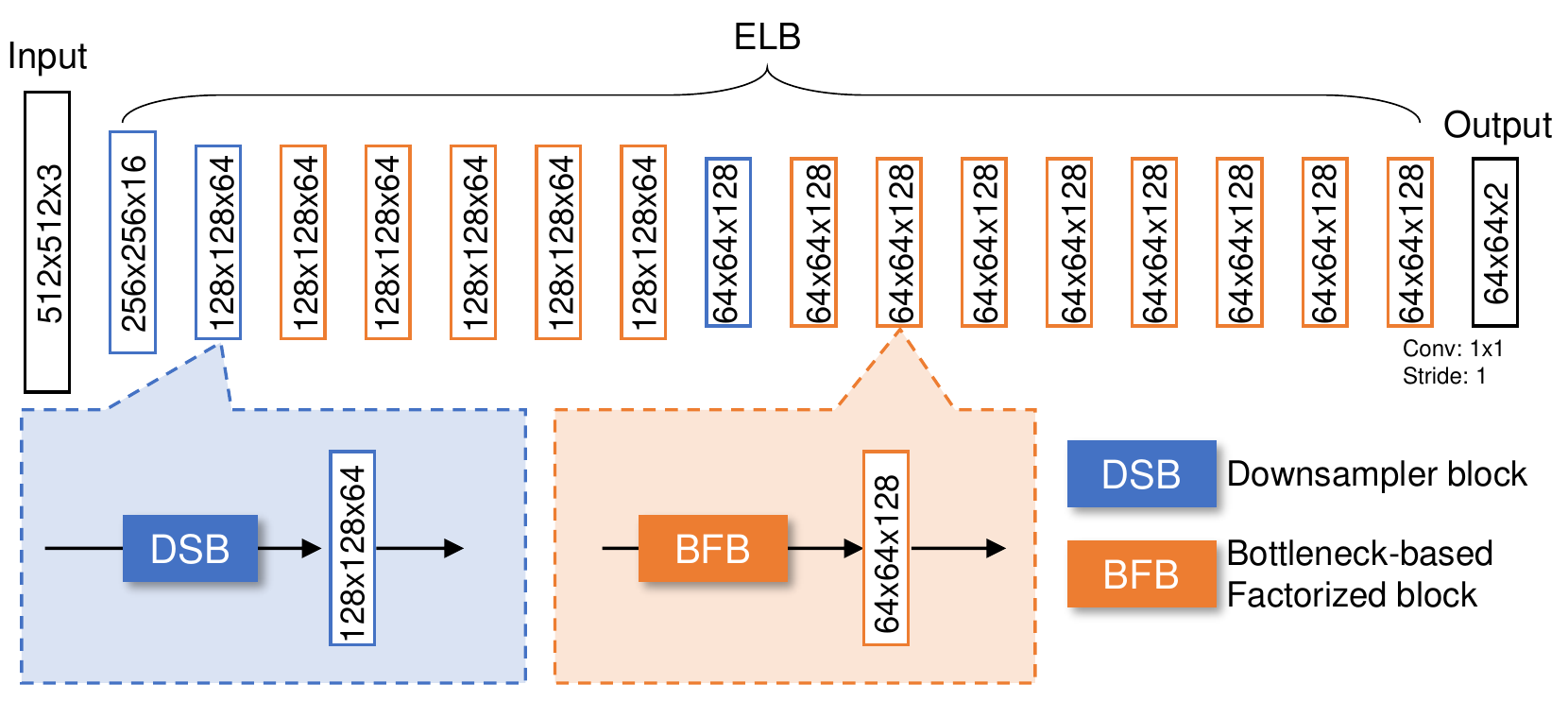}
    \vspace{-0.8cm}
    \caption{\em \small Overview of HLB-based portrait segmentation. The volumes are feature maps generated by layers. Specifically, the blue and orange feature map are generated by downsampler block (DSB) and bottleneck-based factorized block (BFB) respectively. The decoder is a single $1 \times 1$ convolution layer which converts the last feature map to a segmentation map.}
    \label{fig:arch}
    \vspace{-0.5cm}
\end{figure}

Fig.\ref{fig:arch} shows the overall structure of the HLB
architecture for portrait segmentation. It takes input image of $512 \times 512$ pixels and output the mask of segmented portrait. 

The overall structure of HLB architecture is similar to that of \cite{romera2017erfnet}, which starts with two downsampler blocks (DSB) that downsample the feature map by concatenating the output of a $3 \times 3$ convoluational layer with stride 2 and a max-pooling. It is followed by five {\em bottleneck-based factorized blocks} (BFB). After that, another DSB block is used to further downsample the feature map, which is then followed by another set of eight BFBs. 

\vspace{-0.3cm}
\subsection{Bottleneck-based Factorized Blocks}
\label{subsec:factorized-bn}

The {\em bottleneck-based factorized block} (BFB) is the core element of the HLB architecture, which is a special type of residual block \cite{he2016deep}. 

\begin{figure}[t]
    \centering
    \includegraphics[width=0.9\linewidth]{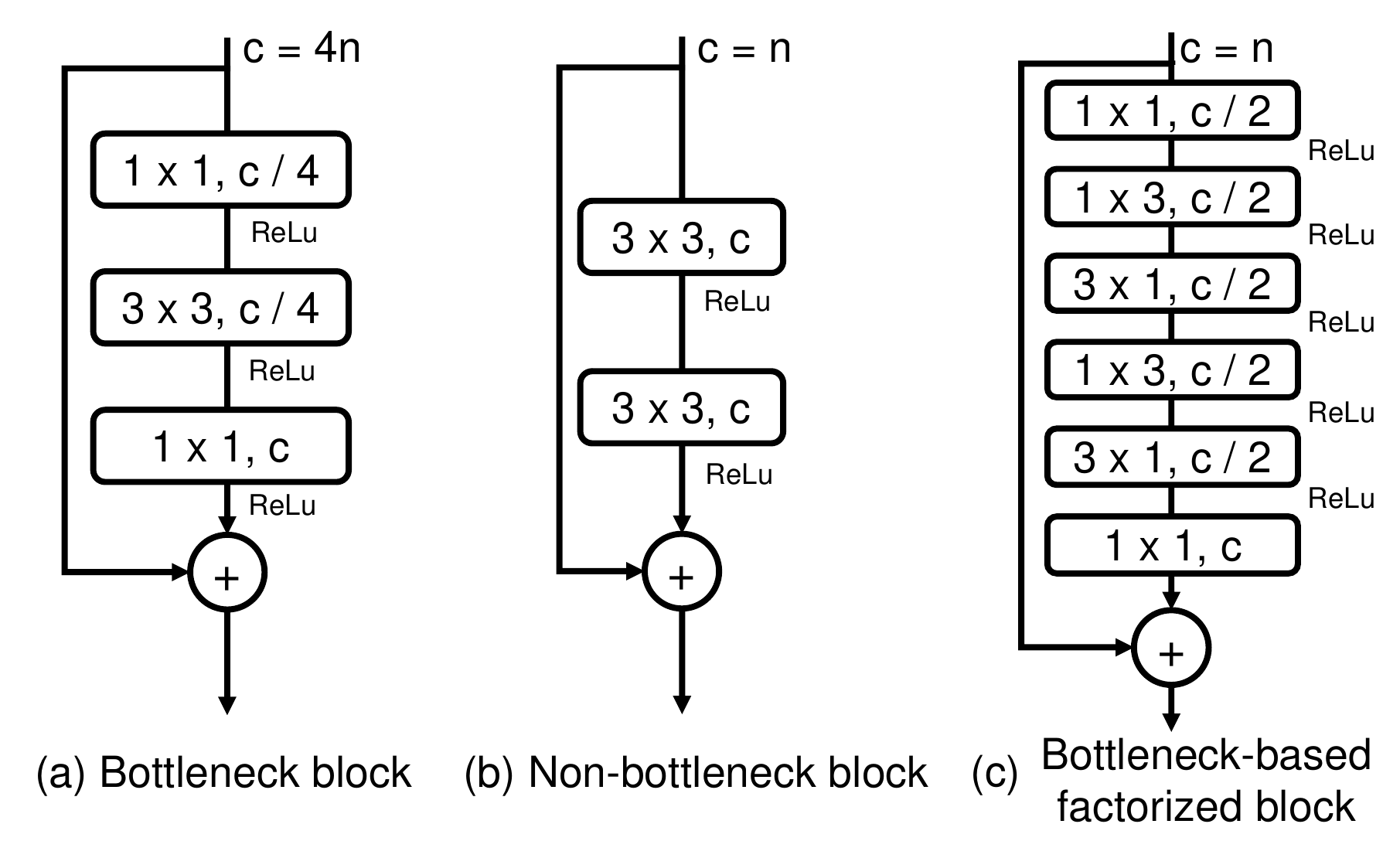}
    \vspace{-0.5cm}
    \caption{\em \small Illustration of two residual blocks originated in \cite{he2016deep}: Bottleneck block (a) and Non-bottleneck block (b), and our proposed Bottleneck-based factorized block (c). ``$c$'' is the channel number of input feature map. ``$n$'' is a constant value for better illustrating the difference between each block. Each box inside the block denotes convolution with kernel size and channel number of output feature map.}
    \vspace{-0.5cm}
    \label{fig:BF}
\end{figure}

\smallskip
\noindent{\bf Residual block}. A residual block can be defined as $$\mathbf{y} = {\cal F}(\mathbf{x}, \{W_i\}) + \mathbf{x},$$ where $\mathbf{x}$ and $\mathbf{y}$ are the input and output feature map, and ${\cal F}(\mathbf{x}, \{W_i\})$ is the residual mapping function implemented as stacked non-linear layers.  There are two different instances of residual block proposed in original work: bottleneck block, which is composed by a bottleneck structure and one $3 \times 3$ convolutional layers (Fig.\ref{fig:BF}(a)) and non-bottleneck block, which is composed by two $3 \times 3$ convoluational layers (Fig.\ref{fig:BF}(b)). As in \cite{he2016deep}, the channel number of input feature map in bottleneck block is four times larger than that in the non-bottleneck block. 

Specifically, the bottleneck structure consists of two $1 \times 1$ convolutional layers added at the beginning and the ending of the block respectively. 
The beginning convoluational layer is implemented by a kernel with size of $1 \times 1 \times c_0 \times c_1$, where $c_0$ and $c_1$ are the number of channels for the input and output feature map, respectively. By setting $c_1 < c_0$, \eg, $c_1 = c_0 / 4$, this layer reduces the width of feature maps in block. The ending convolutional layer is implemented by another kernel with size of $1 \times 1 \times c_1 \times c_0$, which aims to recover the width of feature map back to the original dimension $c_0$. In other words, the size of feature maps inside block is reduced, which forms a bottleneck for the data flow through it. Therefore, despite the channel number of input feature map in bottleneck block is larger than it in non-bottleneck block, they can have similar number of parameters. In terms of performance, the non-bottleneck block usually achieve slightly more accuracy compared to bottleneck block \cite{romera2017erfnet}. 

\smallskip
\noindent{\bf BFB}. The BFB  combines the bottleneck and non-bottleneck blocks together to reduce the number of parameters while retain the similar accuracy.  Specifically in our case, we set $c_1 = c_0 / 2$ instead of $c_1 = c_0 / 4$ for balancing both accuracy and efficiency.  
To further reduce the computation cost, we incorporate the factorized layers \cite{alvarez2016decomposeme} into the residual block. Factorized layers are the decomposition of a standard 2D convolutional layer. Specifically, a 2D convolution layer can be approximately viewed as a linear combination of 1D filters. Let $\mathbf{K} \in \mathbb{R}^{h \times w}$ denote the 2D convolutional kernel\footnote{We omit the channel number of input and output feature map for simplicity.} , where $h$ and $w$ are the height and width of this kernel (usually $h = w$). The 2D kernel $\mathbf{K}$ can be decomposed into two 1D convolutional kernels as $\mathbf{k}^0 \in \mathbb{R}^{1 \times w}, \mathbf{k}^1 \in \mathbb{R}^{h \times 1}$. Let $\mathbf{u}$ and $\mathbf{v}$ denote the input and output feature map, $*$ denote the convolution operation. The 2D convolution can be decomposed as $$\mathbf{v} = \mathbf{K} * \mathbf{u} \rightarrow \mathbf{k}^1 * (\mathbf{k}^0 * \mathbf{u}).$$
As shown in Fig.\ref{fig:BF}(c), we factorize the $3 \times 3$ 2D convolutional kernel into two 1D convolutional kernels with size $1 \times 3$ and $3 \times 1$ respectively. This decomposition further reduces the number of parameters in a standard $3 \times 3$ 2D convolution. Therefore, the new residual block, which we term as {\em bottleneck-based factorized blocks} (BFB) can achieve top performance and meanwhile allowing much better efficiency to satisfy the resource limitation of mobile devices. 

\vspace{-0.4cm}
\subsection{Training}
\label{subsec:loss_function}
\begin{figure}[t]
    \centering
    \includegraphics[width=\linewidth]{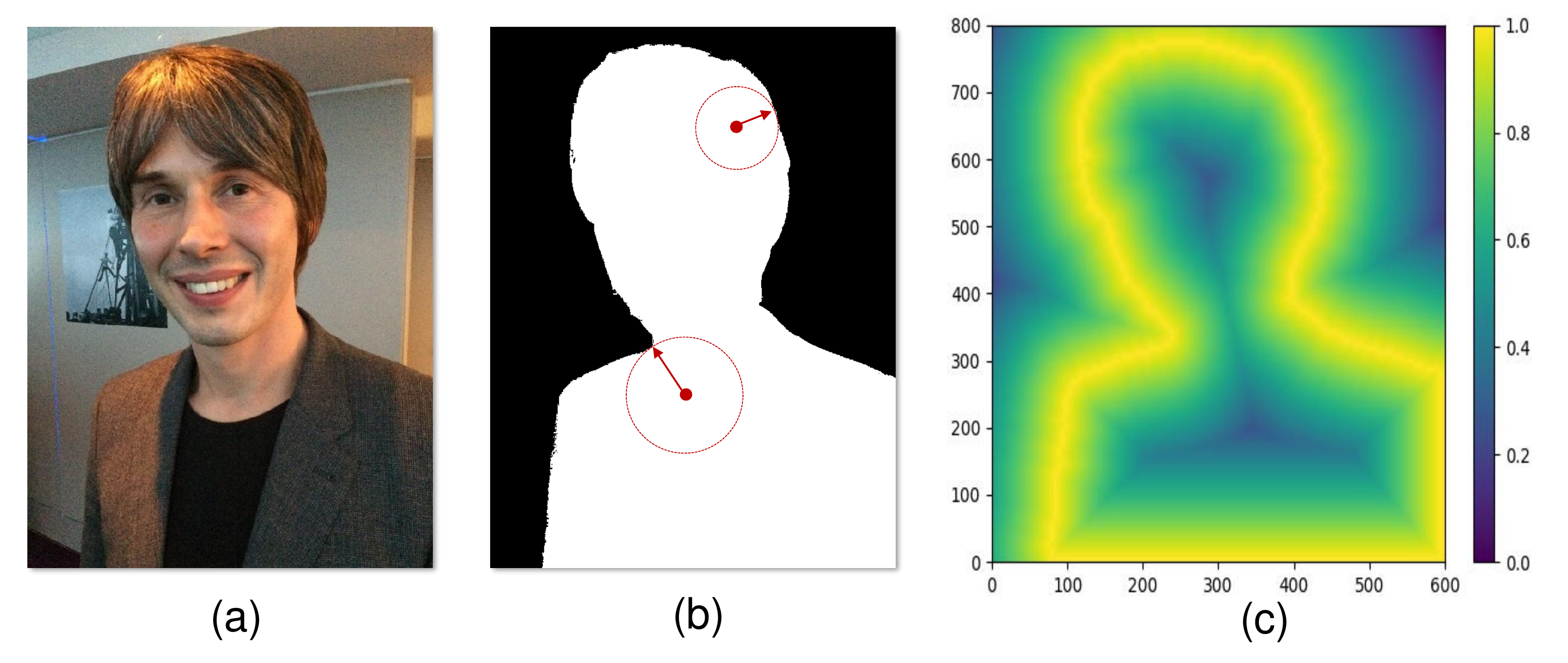}
    \vspace{-0.9cm}
    \caption{\em \small Example of an image (a), weight calculation (b) and corresponding boundary weight map (c). In (b), red point denotes pixel and red circle denote searching range.}
    \label{fig:distance_map}
    \vspace{-0.5cm}
\end{figure}

\vspace{-0.2cm}
For portrait segmentation, it is important to generate accurate boundary in favor of the applications such as background replacement. To this end, we propose an {\em adaptive weight loss} to increase the penalty of the boundary area. Formally, let $\mathbf{m}^p_i$ denote the output of class $p$ at $i$-th pixel of predicted segmentation map and $\hat{p}$ denote the correct class of $i$-th pixel. Let $w_i$ denote the weight of $i$-th pixel. $N$ is the class number ($N=2$ in our task, portrait region or not). $M$ is the number of pixels in an image. The loss function $L$ can be written as
$$L = - \frac{1}{M} \sum_{i=0}^{M} w_i \log\left(\frac{\exp(\mathbf{m}^{\hat{p}}_i)}{\sum_{p=0}^N \exp(\mathbf{m}^p_i)}\right).$$ In general settings \cite{shen2016automatic}, $w_i$ is equal to $1$, which does not highlight the importance of portrait boundary. In our case, we increase the weight of pixels around boundary. Specifically, we first extract the portrait boundary based on the given mask and calculate the closest distance from each pixel to boundary (Fig.\ref{fig:distance_map}(b)). Then we create a boundary weight map by normalizing the distance of each pixel to $[0, 1]$ (Fig.\ref{fig:distance_map}(c)). Let $g_i$ denote the boundary weight of $i$-th pixel in distance map. The $w_i$ is updated to $w_i = 1 + g_i$.   

\vspace{-0.4cm}
\section{Experiments}
\vspace{-0.3cm}

\subsection{Datasets}
\label{sec:dataset}
\begin{figure*}[t]
    \centering
    \vspace{-1cm}
    \includegraphics[width=\linewidth]{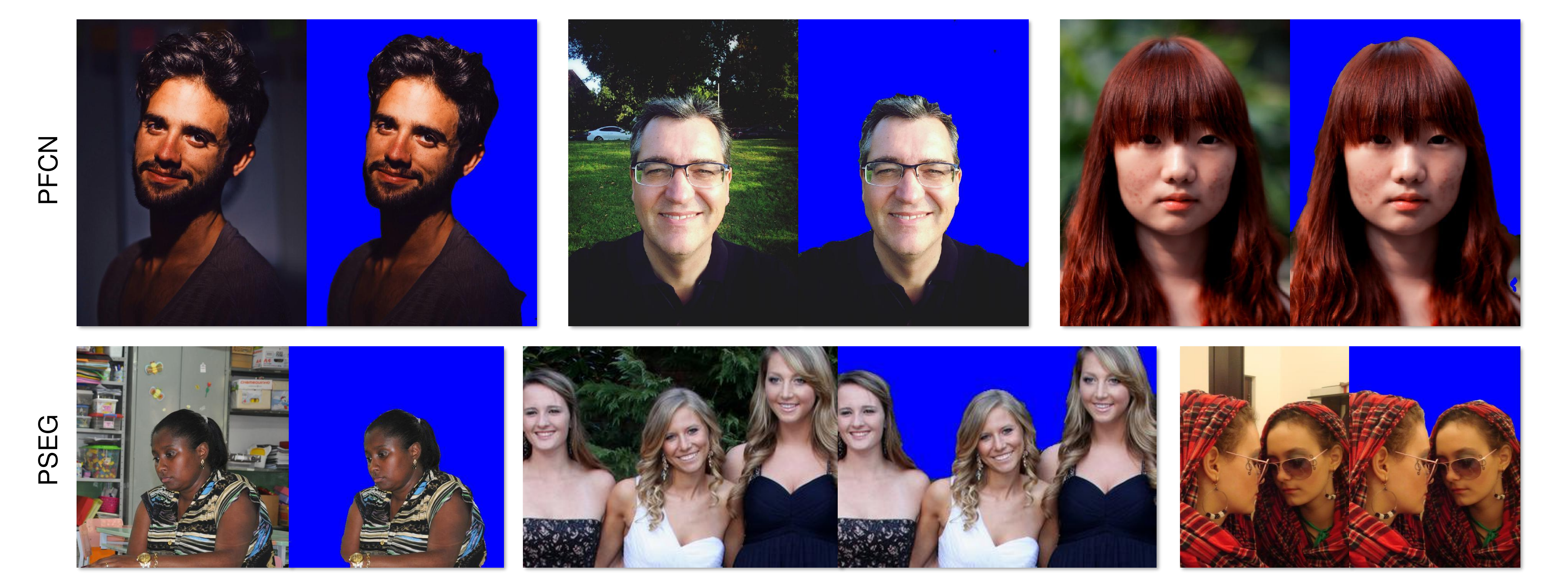}
    \vspace{-1.0cm}
    \caption{\em \small  Example images from  PFCN (top) and PSEG (bottom)  with corresponding segmentation results using our method.}
    \vspace{-0.2cm}
    \label{fig:examples}
\end{figure*}

{\noindent \bf PFCN.} The original PFCN dataset contains $1,800$ portrait images links from Flicker, where $1,500$ images for training and $300$ images for testing. The size of these images is $800 \times 600$ and only one subject is in each image. However, many image links have been expired to date. Only $1,590$ images can be downloaded, which includes $1,277$ images for training and $263$ images for testing. Several examples are shown in Fig.\ref{fig:examples}. 

Due to small scale of this dataset, the methods validated on this dataset may not be well generalized to the real-world application. As such, we create another large scale portrait segmentation dataset (PSEG) as complementary. 

{\noindent \bf PSEG.} We create a new dataset based on Multi-Human Parsing (MHP) dataset \cite{zhao2018understanding}, which provides pixel-wise label for semantically consistent regions of body parts. Based on the annotations of upper body (head and shoulder), the portrait image is cropped around the head region with a randomly bounding box, which can cover the entire head and part of shoulders. In this way, we create $20,514$ portrait images including $18,462$ images for training and $2,052$ images for testing. In contrast to PFCN dataset, which only contains a single subject in view, PSEG dataset contains at least one subject in an image. Several examples are shown in Fig.\ref{fig:examples}.  

\begin{table}[t]
\small
\centering
\resizebox{\linewidth}{!}{
  \begin{tabular}{|c|c|c|c|c|c|c|}
    \hline
    \multirow{2}{*}{Methods}
    & Input size
    & FLOPS 
    & Param. 
    & \multirow{2}{*}{FPS}
    & mIoU \\
    & (w $\times$ h) & (G) & (M) & & $(\%)$ \\
    \hline
    Graph-cut \cite{shi2000normalized} & 600 $\times$ 800 & - & -& -& 80.0$^*$ \\
    \hline
    PFCN+ \cite{shen2016automatic} & 600 $\times$ 800 & - & -& -& 95.9$^*$ \\
    \hline
    PDeepLabv2 \cite{shen2016automatic} & 600 $\times$ 800 & - & -& -& 96.1$^*$ \\
    \hline
    BSN \cite{du2019boundary} & 400 $\times$ 400 & - & -& -& 96.7$^*$ \\
    \hline
    FCN \cite{long2015fully} & 600 $\times$ 800 & - & -& -& 73.1$^*$ \\
    \hline
    \hline
    PIS \cite{zhu2019portrait} & 600 $\times$ 800 & - & -& 23.8 & 95.4$^*$ \\
    \hline
    PortraitNet \cite{zhang2019portraitnet} & 224 $\times$ 224 & 0.5 & 2.1 & 203.2 & 96.6$^*$ \\
    \hline
    \multirow{2}{*}{\bf HLB (ours)} & 224 $\times$ 224 & 0.7 & \multirow{2}{*}{0.6} & 714.3 & 94.6 \\
    \cline{2-3}
    \cline{5-6}
     & 512 $\times$ 512 & 3.8 & & 256.4 & 94.9 \\
    \hline
  \end{tabular}}
  \vspace{-0.4cm}
  \caption{\em \small Quantitative comparison of our method against other existing methods on PFCN dataset. ``*'' denotes the existing methods are trained and validated on original PFCN dataset which can not be accessed now. See text for details.}
  \label{table:compare_efficient}
  \vspace{-0.3cm}
\end{table} 

\subsection{Experimental Settings}
\label{sec:setting}
\noindent{\bf Implementations.} To gather more context in HLB architecture, we use dilated convolutions \cite{yu2015multi} in the second pair of $1 \times 3$ and $3 \times 1$ convolutional layers for the last eight BFBs, with dilated strides of $1, 2, 3, 4, 5, 9, 13, 17$ respectively.
We implement our method using PyTorch 1.0 framework \cite{paszke2017automatic} with CUDA 9.0 on a NIVIDIA 1080ti GPU. The model is trained using Adam optimizer \cite{kingma2014adam} with momentum $0.9$ and weight decay $1e^{-4}$. We use batch size of $16$ and set the starting learning rate to $5e^{-4}$, which is decayed by $0.9$ each epoch. The total epoch number is set to $100$.

\noindent{\bf Data augmentation.} To increase the diversity of training data, we horizontally flip the input image with probability $0.5$ and augment the color by adjusting contrast and brightness with a factor in $[0.8, 1.2]$.

\noindent{\bf Evaluation metric.} The performance is measured using Mean Intersection-over-Union (mIoU) metric, which is defined as $ \frac{1}{N} \sum_{0}^{N} \frac{TP}{TP + FP + FN}$, where $N$ is the class number, TP, FP and FN are number of true positives, false positives and false negatives respectively. 

\vspace{-0.3cm}
\subsection{Comparisons}
\label{sec:results}

We first compare HLB-based portrait segmentation with existing portrait segmentation methods: Graph-cut \cite{shi2000normalized}, PFCN+ \cite{shen2016automatic}, PDeepLabv2 \cite{shen2016automatic}, BSN \cite{du2019boundary}, PIS \cite{zhu2019portrait}, FCN \cite{long2015fully} and PortraitNet \cite{zhang2019portraitnet}. For simplicity, we directly use HLB to denote the HLB-based portrait segmentation. Graph-cut is a graph based method that can extract foreground with user guidance. PFCN+ is adapted from FCN \cite{long2015fully} trained on PFCN dataset. PDeepLabv2 and BSN are both built on DeepLabv2 ResNet101 model for portrait segmentation. FCN method denotes directly using the model trained on semantic segmentation for portrait segmentation using `person' class. PIS and PortraitNet both adopt MobileNetv2 as the backbone architecture for light weight design. 
Note that all of these existing methods are trained and validated on original PFCN dataset, which can not be accessed now (see Sec.\ref{sec:dataset}). However, the proposed HLB can only be trained on the current dataset, which is only a subset of original one. Since the code implementations of these existing methods are not realised, we can hardly re-train these methods on the current PFCN dataset. Thus it is difficult to fairly compare HLB with these existing methods. Table \ref{table:compare_efficient} shows the performance of each existing method on original PFCN dataset and HLB on current dataset \footnote{The performance of Graph-cut, PFCN+, PDeepLabv2 are referred from \cite{du2019boundary}, FCN is referred from \cite{zhu2019portrait}.}. ``FLOPS'' denotes the floating-point operations per second. ``Param.'' stands for the number of parameters. Despite the dataset used for evaluation is not exactly same, we highlight that the accuracy performance of proposed HLB is on par with the state-of-the-arts, yet achieves much better running efficiency. 

For the model efficiency analysis, only PIS and PortraitNet are built based on light weight architecture. They test the inference time on the same type of GPU (NVIDIA 1080ti) with us, which can achieve $23.8$ FPS on input size $600 \times 800$ and $203.2$ FPS on input size $224 \times 224$ respectively. Compared to these two methods, the HLB-based portrait segmentation has fewer parameters and runs faster, which achieves $714.3$ FPS and $256.4$ FPS on input size $224 \times 224$ and $512 \times 512$ respectively.

\begin{table}[t]
\small
\centering
\resizebox{\linewidth}{!}{
  \begin{tabular}{|c|c|c|c|c|c|}
    \hline
    \multirow{2}{*}{Methods}
    & \multirow{2}{*}{FLOPS (G)}
    & \multirow{2}{*}{Param. (M)}
    & \multirow{2}{*}{FPS}
    & \multicolumn{2}{c|}{mIoU $(\%)$} \\
    \cline{5-6}
    & & & & PFCN & PSEG \\
    \hline
    ERFNet \cite{romera2017erfnet} & 13.4 & 2.1 & 128.2 & 94.6 & 93.9 \\
    \hline
    ENet \cite{paszke2016enet} & 2.1 & 0.4 & 220.7 & 94.2 & 93.9 \\
    \hline
    ICNet \cite{zhao2018icnet} & 3.5 & 7.7 & 83.3 & 87.8 & 90.0 \\
    \hline
    BiSeNet \cite{yu2018bisenet} & 14.5 & 12.8 & 164.0 & 94.5 & 94.4 \\
    \hline
    \hline
    {\bf HLB (ours)} & 3.8 & 0.6 & 256.4 & 94.9 & 94.0 \\
    \hline
  \end{tabular}}
  \vspace{-0.4cm}
  \caption{\em \small Comparisons of our method with other efficient networks used in semantic segmentation.}
  \label{table:compare_semantic}
  \vspace{-0.5cm}
\end{table} 

Furthermore, we compare HLB with existing efficient networks used in semantic segmentation: ERFNet \cite{romera2017erfnet}, ENet \cite{paszke2016enet}, ICNet \cite{zhao2018icnet} and BiSeNet \cite{yu2018bisenet}. The backbone of BiSeNet is ResNet18. We re-train these networks on PFCN and PSEG datasets respectively using their default training settings on the same environment as ours. The input image for each method is set to $512 \times 512$, the same as ours. The accuracy and efficiency analysis is reported in Table \ref{table:compare_semantic}. The results reveal that our HLB-based portrait segmentation method can run $256.4$ FPS, which is faster than other methods. Note the ENet has less FLOPS and Parameters than ours, yet it is slower than our method. This is probably due to the ENet has much more skip connections than ours, which is slower under the implementation of PyTorch.  At the same time, our method can achieve the best accuracy performance ($94.9\%$) on current PFCN dataset and competitive results ($94.0\%$) on PSEG dataset.  Fig.\ref{fig:examples} shows several segmentation resutls of our method on two datasets.

\vspace{-0.4cm}
\subsection{Ablation Study} 
\vspace{-0.2cm}
We investigate the effect of the channel decreasing rate in bottleneck of BFB. In this experiment, we enlarge the channel decreasing rate from 2 to 4. In this way, the width of block will be further reduced. Table \ref{table:ablation} shows the quantitative comparison of BFB with channel decreasing rate 4 and 2 on PFCN dataset. From the results we can see by changing the channel decreasing rate to 4, the FLOPS, number of parameters are notably reduced, which lead to an improvement in FPS from $256.4$ to $370.4$. However, the accuracy is degraded notably by $1.2\%$. Considering the trade-off between accuracy and efficiency, we employ the channel decreasing rate 2 in our setting.

\begin{table}[t]
\small
\centering
\resizebox{\linewidth}{!}{
  \begin{tabular}{|c|c|c|c|c|}
    \hline
    Method& FLOPS (G) &  Param. (M) & FPS & mIoU ($\%$) \\
	\hline
	BFB-DR4 & 1.4 & 0.2 & 370.4 & 93.7 \\
	\hline
	BFB-DR2 & 3.8 & 0.6 & 256.4 & 94.9 \\
    \hline
  \end{tabular}}
  \vspace{-0.4cm}
  \caption{\em \small Comparisons of bottleneck-based factorized blocks with different channel decreasing rate. ``DR4'' and ``DR2'' indicate the channel decreasing rate is 4 and 2 respectively.}
  \label{table:ablation}
  \vspace{-0.5cm}
\end{table}

\vspace{-0.6cm}
\section{Conclusion}
\vspace{-0.3cm}
In this paper, we propose an highly light-weight backbone (HLB) architecture for portrait segmentation. The proposed architecture runs much faster than existing methods, yet achieves competitive performance on several datasets. The core element of HLB is a novel bottleneck-based factorized block (BFB), which is the combination of bottleneck structure with factorized layers that can greatly reduce the number of parameters while retaining the good learning capacity. The experiments performed on PFCN and PSEG dataset demonstrates the effectiveness of HLB architecture on portrait segmentation task.

\bibliographystyle{IEEEbib}
\bibliography{strings}

\end{document}